\documentclass[times, 10pt,twocolumn]{article} 
\usepackage{latex8}
\usepackage{times}

\usepackage{epsfig}
\usepackage{graphics}

\usepackage{amssymb}

\usepackage[latin1]{inputenc} 

\begin{document}

\title{Automatic Face Recognition System Based on Local Fourier-Bessel Features}

\author{Yossi Zana, Roberto M. Cesar-Jr and Regis de A. Barbosa\\
Instituto de Matemática e Estatística - USP\\
R. do Matão, 1010 - Cidade Universitária\\
CEP: 05508-090, São Paulo - SP, Brazil\\
\{zana,cesar\}@vision.ime.usp.br, regisb@ime.usp.br}

\maketitle
\thispagestyle{empty}

\begin{abstract}
  We present an automatic face verification system inspired by known
  properties of biological systems. In the proposed algorithm the
  whole image is converted from the spatial to polar frequency domain
  by a Fourier-Bessel Transform (FBT). Using the whole image is
  compared to the case where only face image regions (local analysis)
  are considered. The resulting representations are embedded in a
  dissimilarity space, where each image is represented by its distance
  to all the other images, and a Pseudo-Fisher discriminator is built.
  Verification test results on the FERET database showed that the
  local-based algorithm outperforms the global-FBT version. The
  local-FBT algorithm performed as state-of-the-art methods under
  different testing conditions, indicating that the proposed system is
  highly robust for expression, age, and illumination variations. We
  also evaluated the performance of the proposed system under strong
  occlusion conditions and found that it is highly robust for up to
  50\% of face occlusion. Finally, we automated completely the
  verification system by implementing face and eye detection
  algorithms. Under this condition, the local approach was only
  slightly superior to the global approach.
\end{abstract}

\section{Introduction}

Face verification and recognition tasks are highly complex due to the
many possible variations of the same subject in different conditions,
like illumination, facial expression, and age. Many developers of face
recognition algorithms adopted a biologically inspired approach in
solving these problems (for a review, see \cite{Calderetal2001}), thus
contributing both to understand human face processing and to build
efficient face recognition technologies.

The approach described in the present paper was inspired by
developments in neurophysiology and cognitive psychology and its
fundamentals were first described by \cite{ZanaCesar2005}. It is based
on image representation that may be analogous to those used by the
human visual system (HVS). In particular, we evaluated the performance
of a face verification system whose primary features were the
magnitude of radial and angular components of faces images, and
representation in a dissimilarity space. The main contribution of this
paper is the proposal to use local analysis approach, in contrast to
the previously used global analysis approach\footnote{Partial results based on a preliminary version of the system were submitted in \cite{Zanaetal2005s}}. We show that a system
based on the new method achieves state-of-the-art performance level.
Moreover, we demonstrate that the proposed system is robust to typical
variations in face images, like facial expression, age, illumination,
and partial occlusion.

The paper is organized as follows: in the next section, we briefly
introduce the reader to the primary spatial processing by the HVS and
to the related literature. We describe in section 3 the Fourier-Bessel
Transform (FBT) and the proposed algorithm in section 4. We introduce
the face database and testing methods in Section 5. The experimental
results are presented in Section 6 and in the last section we discuss
the results and ongoing work.

\section{Background and previous work}

Most of the current face recognition and verification algorithms are
based on feature extraction from a Cartesian perspective, typical to
most analog and digital imaging systems. On the other hand, the HVS is
known to process visual stimuli by fundamental shapes defined in polar
coordinates. In the early stages the visual image is filtered by
neurons tuned to specific spatial frequencies and location in a linear
manner \cite{DevaloisDevalois1990}. In further stages, these neurons
output is processed to extract global and more complex shape
information, such as faces \cite{Perretetal1982}.  Electrophysiological
experiments in monkey's visual cerebral areas showed that the
fundamental patterns for global shape analysis are defined in polar
and hyperbolic coordinates \cite{Gallantetal1993}.  Global pooling of
orientation information was also shown by psychophysical experiments
to be responsible for the detection of angular and radial Glass dot
patterns \cite{WilsonWilkinson1998}.  Thus, it is evident that
information regarding the global polar content of images is
effectively extracted by and is available to the HVS. In
\cite{ZanaCesar2005} we introduced the representation of face images
in the polar frequency domain by global two-dimensional FBT features.
However, one of the disadvantages of global feature extractions is the
rough representation of peripheral regions.  The HVS compensates this
effect by eye saccades, moving the fovea from one point to the other
in the scene. Here we propose to apply the FBT at strategic regions
\cite{HeiseleKoshizen2004}, namely the eyes region. Moreover, we also
integrated face and eyes detection algorithms, which makes the
verification system completely automatic.

\section{Fourier-Bessel analysis}

The FB series \cite{Foxetal2003} is useful to describe the radial and
angular components in images. FBT analysis starts by converting the
coordinates of a region of interest from Cartesian $\left( {x,y}
\right)$ to polar $\left( {r,\theta } \right)$. The $f\left( {r,\theta
  } \right)$ function is represented by the two-dimensional FB series,
defined as
\\
$$
f(r,\theta )=\sum\limits_{i=1}^\infty {\sum\limits_{n=0}^\infty
  {A_{n,i}Jn(\alpha _{n,i}r)\cos (n\theta )} }\\
$$
\begin{equation}
+\sum\limits_{i=1}^\infty {\sum\limits_{n=0}^\infty {Bn,iJn(\alpha 
_{n,i}r)\sin (n\theta )} } 
\end{equation}
where $J_n$ is the Bessel function of order $n$, $f(R,\theta )=0$ and
$0\le r\le R. \quad \alpha _{n,i}$ is the $i{th}$ root of the $J_n$
function, i.e. the zero crossing value satisfying $J_n(\alpha
_{n,i})=0$. $R$ is the radial distance to the edge of the image. The
orthogonal coefficients $A_{n,i}$ and $B_{n,i}$ are given by

$$
A_{0,i}=\frac{1}{\pi R^2J^2_1(\alpha _{n,i})}\int\limits_{\theta
  =0}^{\theta =2\pi } {\int\limits_{r=0}^{r=R} {f(r,\theta )rJ_n}}
$$
\begin{equation}
  (\frac{\alpha n,i}{R}r)drd\theta 
\end{equation}

if $B_{0,i}=0$ and $n=0$;
\\
\\
$$
\left[ {\begin{array}{l}
      A_{n,i} \\
      B_{n,i} \\
      \end{array}} \right]=\frac{2}{\pi 
    R^2J^2_{n+1}(\alpha _{n,i})}\\
    \int\limits_{\theta =0}^{\theta =2\pi } {\int\limits_{r=0}^{r=R} {f(r,\theta     )rJ_n}}
$$
\begin{equation}
  (\frac{\alpha _{n,i}}{R}r) \left[ {\begin{array}{l}
        \cos (n\theta ) \\ 
        \sin (n\theta ) \\ 
     \end{array}} \right]drd\theta
\end{equation}

if $n>0.$

\section{Face verification using FBT}

The proposed algorithm is based on image registration and
normalization, and two subsequent feature extraction steps followed by
a classifier formation. After the first two steps, we extract the FB
coefficients from the images, we compute the pairwise Cartesian distance
between all the FBT-representations and re-define each object by its
distance to all other objects. In the last stage we train a pseudo
Fisher classifier.  We tested this algorithm on the whole image
(global analysis) and on the combination of three facial regions
(local analysis).

\subsection{Face registration and normalization}
Face representations requires prior image registration and usually a
spatial and luminance normalization pre-processing. Assuming the
sample images contain a single face, we detected the head with a
cascade of classifiers \cite{ViolaJones2001} and estimated the
location of the eyes region with an Active Appearance Model algorithm
\cite{Cootesetal2001}.  Within this region, we used flow field
information \cite{KothariMitchell1996} to determine the eyes center.
Using the eyes coordinates, we translated, rotated, and scaled the
images so that the eyes were registered at specific pixels. Next, the
images were cropped to 130 x 150 pixels size and a mask was applied to
remove most of the hair and background. The unmasked region was
histogram equalized and normalized to zero mean and a unitary standard
deviation.

\subsection{Spatial to polar frequency domain}
\begin{figure}[ht] 
  \begin{center}
  \begin{tabular}{cccc}
     
      Normalized &
      Right eye &
      Between eyes &
      Left eye \\
      \epsfig{file=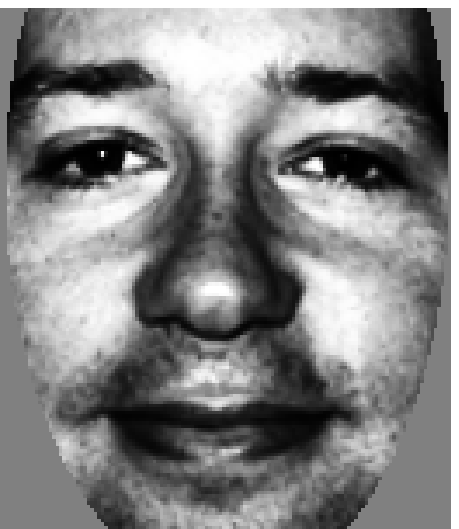, width=0.085\textwidth} &
      \epsfig{file=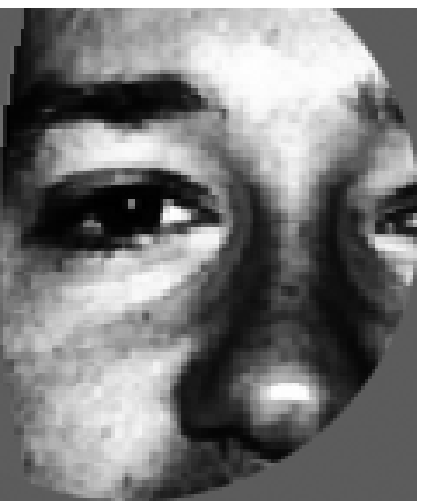, width=0.085\textwidth} &
      \epsfig{file=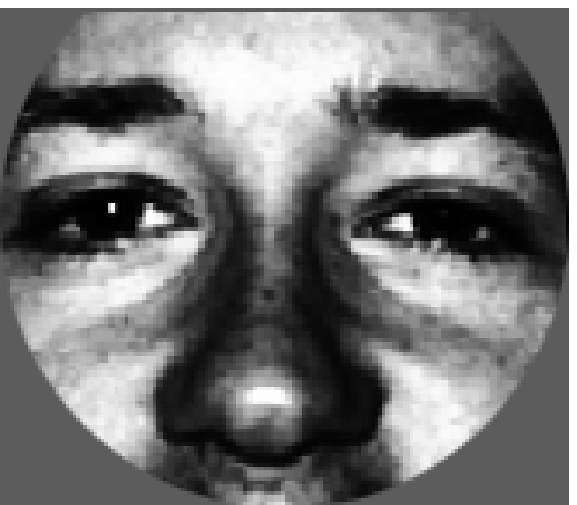, width=0.085\textwidth} &
      \epsfig{file=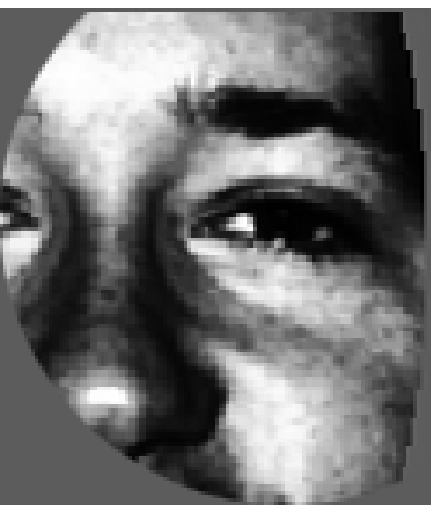, width=0.085\textwidth}

    \end{tabular}
  \caption{Sample of a normalized whole face image and the regions that were used for the local analysis}
  \label{fig:examples-normal}
  \end{center}
\end{figure}

Images were transformed by a FBT up to the 30$^{th}$ Bessel order and
6$^{th}$ root with angular resolution of 3\r{ }, thus obtaining 372
coefficients. These coefficients correspond to a frequency range of up
to 30 and 3 cycles/image of angular and radial frequency,
respectively. This frequency range was selected based on earlier tests
\cite{ZanaCesar2005} with the small-size ORL face database
\cite{SamariaHarter1994}. We tested the FBT descriptors of the whole
image, as well as a combination of the upper right region, upper
middle region, and the upper left region (Figure
\ref{fig:examples-normal}).

\subsection{Polar frequency to dissimilarity domain}
We built a symmetric dissimilarity matrix $D\left( {{\rm {\bf t}},{\rm {\bf t}}}
\right)$ defined as the Euclidean distance between all training FBT
images ${\rm {\bf t}}$. In this space, each object is represented by
its dissimilarity to all objects. This approach is based on the
assumption that the dissimilarities of similar objects to ``other
ones'' is about the same \cite{Duinetal1997}.  Among other advantages
of this representation space, by fixing the number of features to the
number of objects, it avoids a well known phenomenon, where
recognition performance is degraded as a consequence of small number
of training samples as compared to the number of features.

\subsection{Classifier}
Test images were classified based on a pseudo FLD using a two-class
approach. A FLD is obtained by maximizing the (between subjects
variation)/(within subjects variation) ratio \cite{Fukunaga1990}. Here
we used a minimum-square error classifier implementation
\cite{ScurichinaDuin1996}, which is equivalent to the FLD for
two-class problems \cite{Fukunaga1990}. In these cases, after shifting
the data such that it has zero mean, the FLD can be defined as
\begin{equation}
  g\left( {\bf x} \right)=\left[ {D\left( {{\rm 
            {\bf t}},{\bf x}} \right)-\frac{1}{2}\left( {{\rm {\bf
              m}}_1 -{\rm {\bf m}}_2 }  
      \right)} \right]^T{\rm {\bf S}}^{-1}\left( {{\rm {\bf m}}_1
      -{\rm {\bf m}}_2  
    } \right)
\end{equation}
where ${\bf x}$ is a probe image, ${\rm {\bf S}}$ is the pooled
covariance matrix, and ${\rm {\bf m}}_i$ stands for the mean of class
$i$. The probe image ${\bf x}$ is classified as corresponding to
class-1 if $g({\bf x}) \ge 0$ and to class-2 otherwise.  However, as
the number of training objects and dimensions is the same in the
dissimilarity space, the sample estimation of the covariance matrix
${\rm {\bf S}}$ becomes singular, and the classifier cannot be built.
One solution to the problem is to use a pseudo-inverse and augmented
vectors \cite{ScurichinaDuin1996}. Thus, Eq. 6 is replaced by
\begin{equation}
  \label{eq1}
  g\left( {\bf x} \right)=\left( {D\left( {{\rm 
            {\bf t}},{\bf x}} \right),1} \right)\left( {D\left( {{\bf
            t},{\bf t}} \right),I}  
  \right)^{\left( {-1} \right)}
 \end{equation}
 where $\left( {D\left( {{\rm {\bf t}},{\bf x}} \right),1} \right)$ is
 the augmented vector to be classified and $\left( {D\left( {{\bf
           t},{\bf t}} \right),I} \right)$ is the augmented training
 set. The inverse $\left( {D\left( {{\bf t},{\bf t}} \right),I}
 \right)^{\left( {-1} \right)}$ is the Moore-Penrose Pseudo-inverse
 which gives the minimum norm solution. The current $L$-classes
 problem can be reduced and solved by the two-classes solution
 described above. The training set was split into $L$ pairs of
 subsets, each pair consisting of one subset with images from a single
 subject and a second subset formed from all the other images.  A
 pseudo-FLD was built for each pair of subsets. A probe image was
 tested on all $L$ discriminant functions, and a ``posterior
 probability'' score was generated based on the inverse of the
 Euclidean distance to each subject.

\begin{figure*}[th]
  \begin{center}
  \begin{tabular}{cc}

      Age (1-34 months) &
      Age (18-34 months) \\
      \epsfig{file=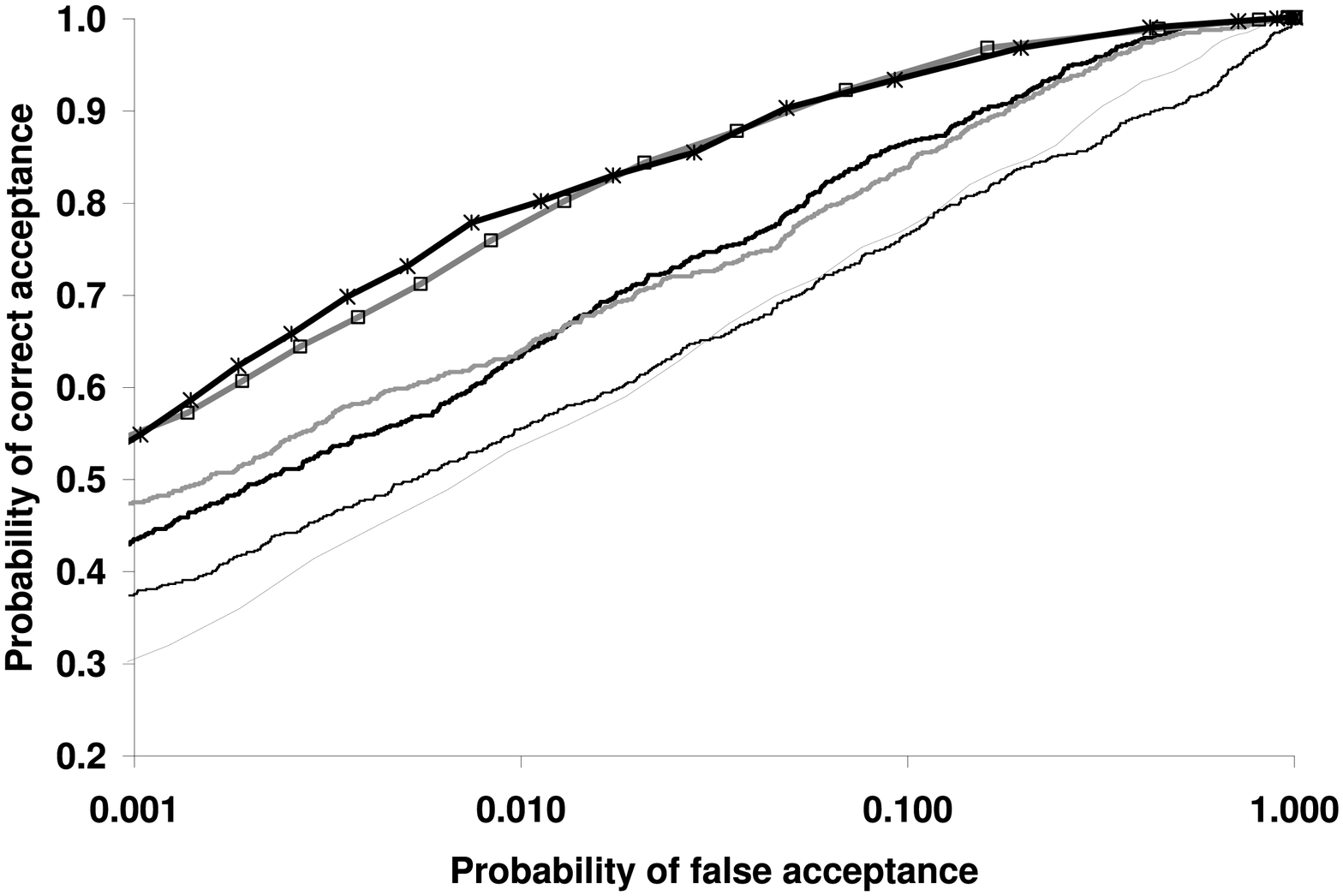, width=0.45\textwidth} &
      \epsfig{file=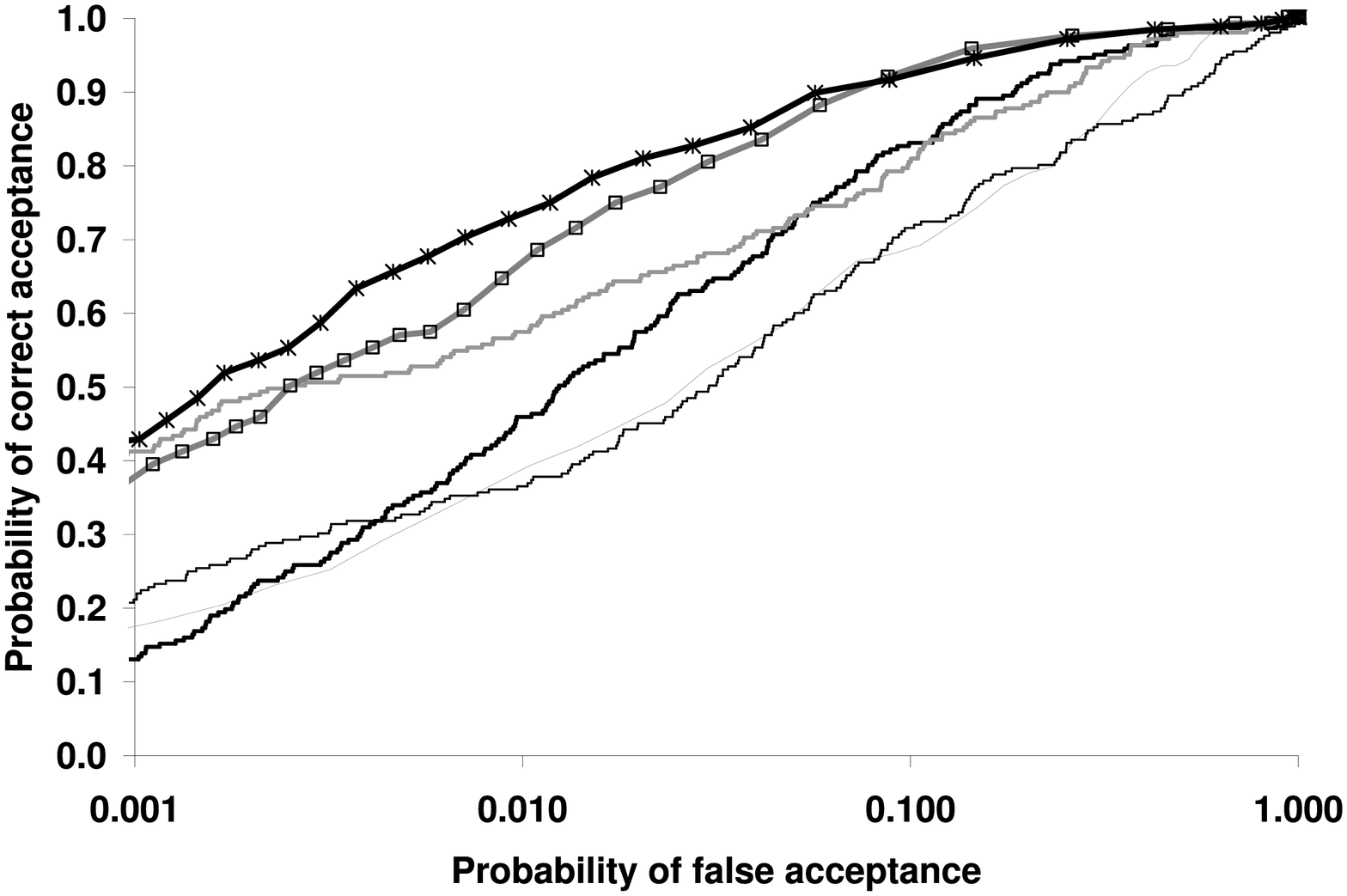, width=0.45\textwidth} \\

      Expression &
      Illumination \\
      \epsfig{file=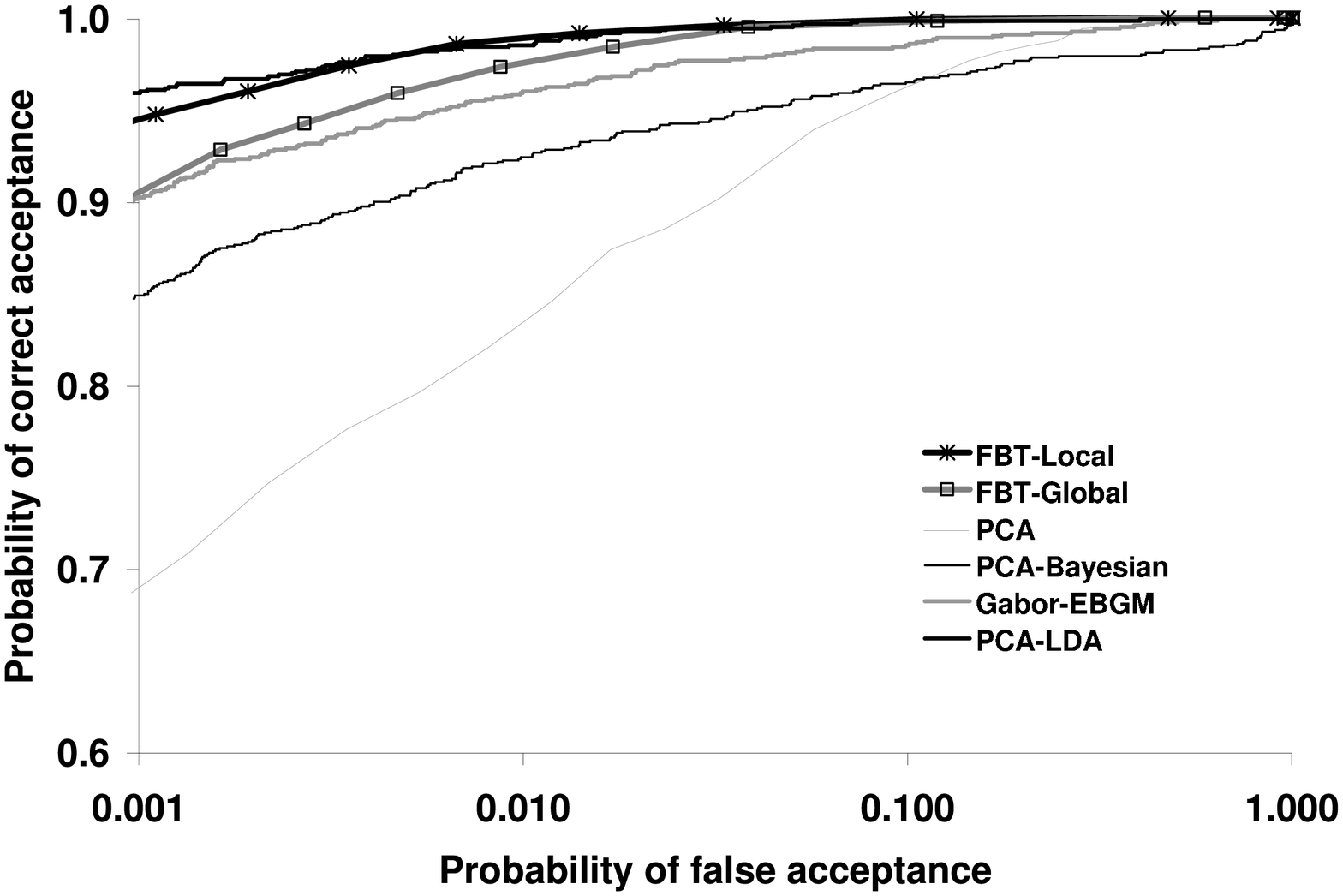, width=0.45\textwidth} &
      \epsfig{file=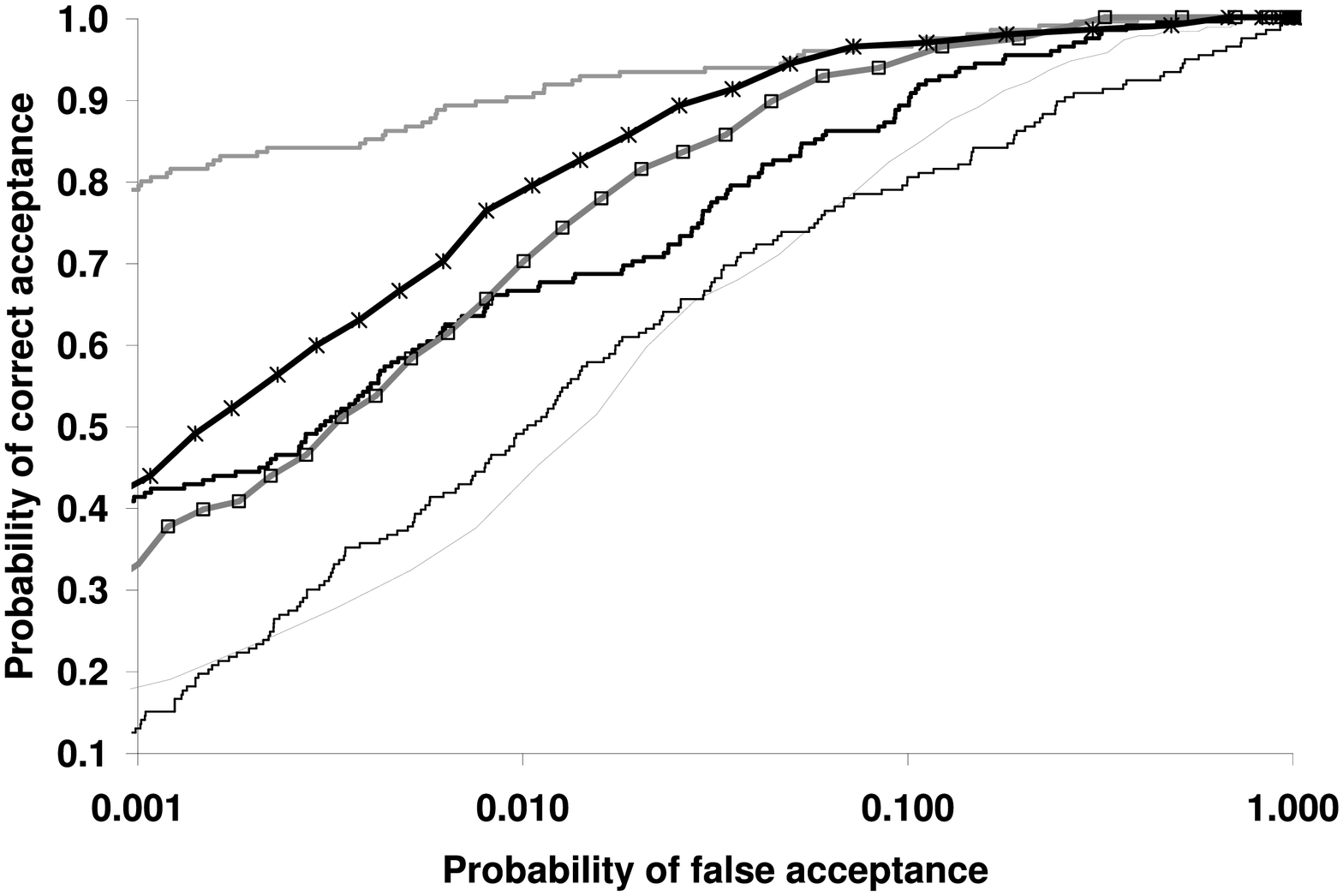, width=0.45\textwidth} \\

  \end{tabular}
  \caption{ROC functions of the FBT, PCA, and previous algorithms on the age, expression and illumination subsets.}
  \label{fig:roc-fbt}
  \end{center}
\end{figure*}

\section{Database, preprocessing, and testing procedures}

We used the FERET database, due to its large number of individuals and
rigid testing protocols that allow precise performance comparisons
between different algorithms \cite{Phillipsetal1998}. Here we compare
our algorithm performance with a ``baseline'' algorithm and with the
published results of three successful approaches \cite{Rizvietal1998}.
As a baseline algorithm we implemented a standard PCA-based algorithm
\cite{TurkPentland1991}.  The principal components were based on a set
of 700 images selected randomly from the gallery subset. Not all 1196
images were used, due to the huge RAM memory that such operation
requires. The first three principal components, that encode basically
illumination variations \cite{Hallinan1994}, were excluded before
projecting of the training and test images. The three other approaches
are: Gabor wavelets combined with elastic bunch graph matching (EBGM)
\cite{Wiskottetal1997}, localized facial features extraction followed
by a Linear Discriminant Analysis (LDA) \cite{EtemadChellappa1997},
and a Bayesian generalization of the LDA method
\cite{Moghaddametal2000}.

In the FERET protocol, a \emph{gallery} set of one frontal view image
from 1196 subjects is used to train the algorithm and a different
dataset is used as probe. All images are gray-scale 256 x 384 pixels
size. We used the four probe sets, termed \emph{FB}, \emph{DupI},
\emph{DupII} and \emph{FC}. The FB dataset is constituted of a single
image from 1195 subjects, taken from the same subjects in the gallery
set, after an interval of a few seconds, but with a different facial
expression. The DupI and DupII datasets include 722 or 234 images,
respectively. The DupI images were taken immediately or up to 34 months
after the gallery images, while the images in DupII were taken at
least 18 months after the gallery images. The FC subset contains 194
images of subjects under different lighting conditions.

The eyes coordinates were extracted automatically, as described in
Section 4.1. Approximately 1\% of the faces were not localized, in
which cases the eyes region coordinates were set to a fix value
derived from the mean of the located faces.  The final mean error was
3.6 $\pm$ 5.1 pixels. In order to estimate the system performance
under minimal localization errors, we executed a second series of
experiments in which ground-truth information was used. The face
registration was followed by a normalization step, as described in
Section 4.1. The same pre-processing procedure was used in previous
algorithms, except for the Gabor-EBGM system where a special
normalization procedure was used.

The performance of the system was evaluated by verification tests
according to the FERET protocol \cite{Phillipsetal1998}.  Given a
gallery image $g$ and a probe image $p$, the algorithm verifies the
claim that both were taken from the same subject. The verification
probability $P_{V}$ is the probability of the algorithm accepting the
claim when it is true, and the false-alarm rate $P_{F}$ is the
probability of incorrectly accepting a false claim. The algorithm
decision depends on the posterior probability score $si\left( k
\right)$ given to each match, and on a threshold $c$. Thus, a claim is
confirmed if $si\left( k \right)\le c$ and rejected otherwise. A plot
of all the combinations of $P_{V}$ and $P_{F}$ as a function of $c$ is
known as a receiver operating characteristic (ROC). $P_{V}$ and
$P_{F}$ were calculated as the number of confirmations divided by the
number of correct or incorrect matches, respectively. This procedure
was repeated for 100 equally spaced threshold levels. Training and
tests were done with the PRTools toolbox \cite{Duin2000}.

\section{Results}

\subsection{Semi-automatic system}
Figure \ref{fig:roc-fbt} shows the performance of the proposed
algorithm in the verification test. On the expression dataset the
global and local FBT versions performed at about the same level as the
previous best and second-best algorithms, respectively. On both age
datasets the FBT algorithms outperformed the previous algorithms, with
the local version being slightly superior. On the illumination dataset
the global and local FBT algorithms were equal or better than the
second-best previous algorithm (PCA+LDA).

\subsection{Partial occlusion}
\begin{figure}[h!]
  \begin{center}
  \begin{tabular}{ccc}

      No occlusion &
      Eye + mouth &
      Mouth + nose \\
      \epsfig{file=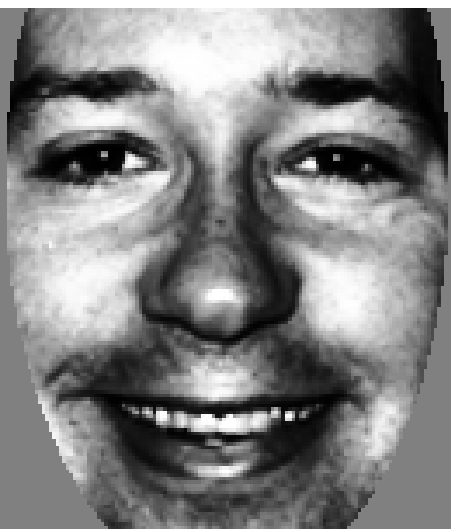, width=0.13\textwidth} &
      \epsfig{file=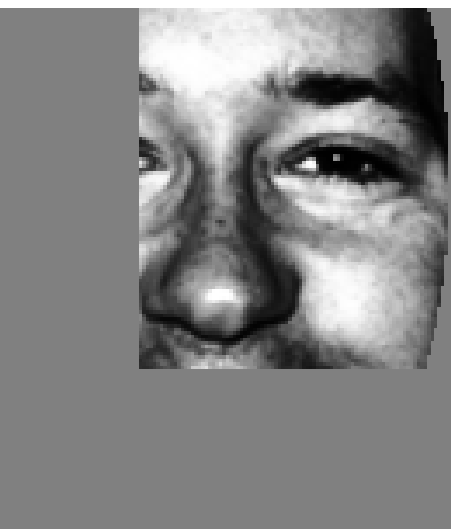, width=0.13\textwidth} &
      \epsfig{file=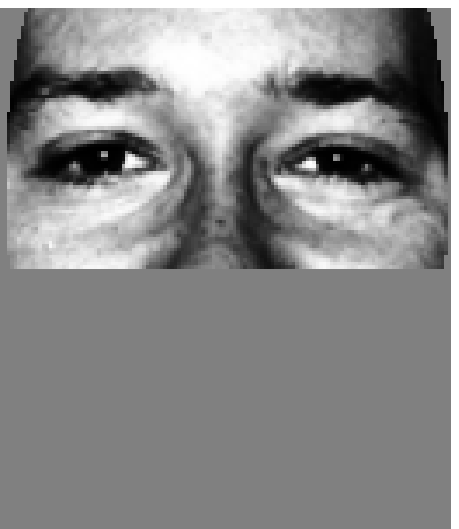, width=0.13\textwidth} \\

  \end{tabular}
  \caption{Examples of image occlusion.}
  \label{fig:examples-occlusion}
  \end{center}
\end{figure}

\begin{figure*}[htb!]
  \begin{center}
  \begin{tabular}{cc}

      Age (1-34 months) &
      Age (18-34 months) \\
      \epsfig{file=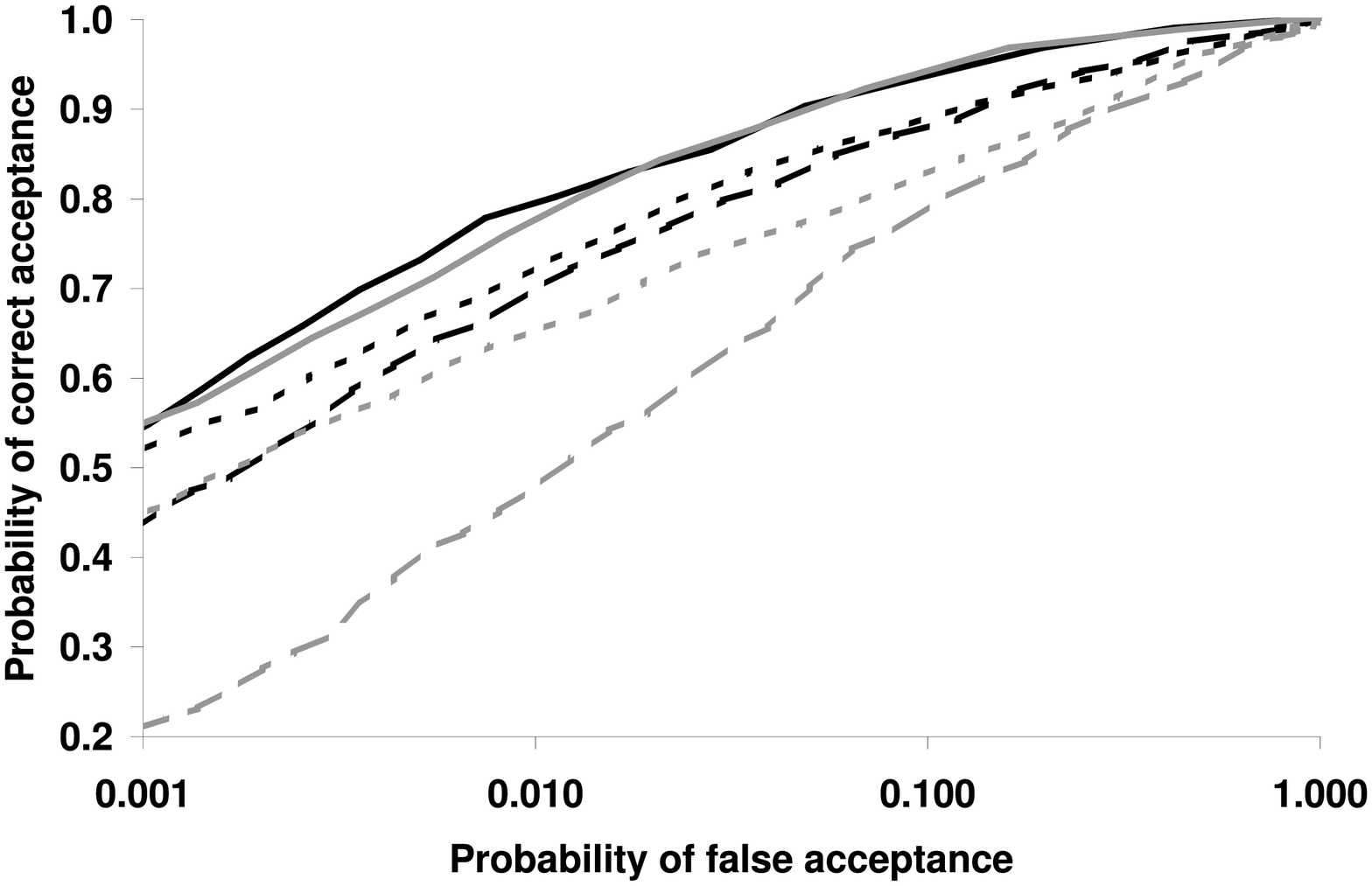, width=0.45\textwidth} &
      \epsfig{file=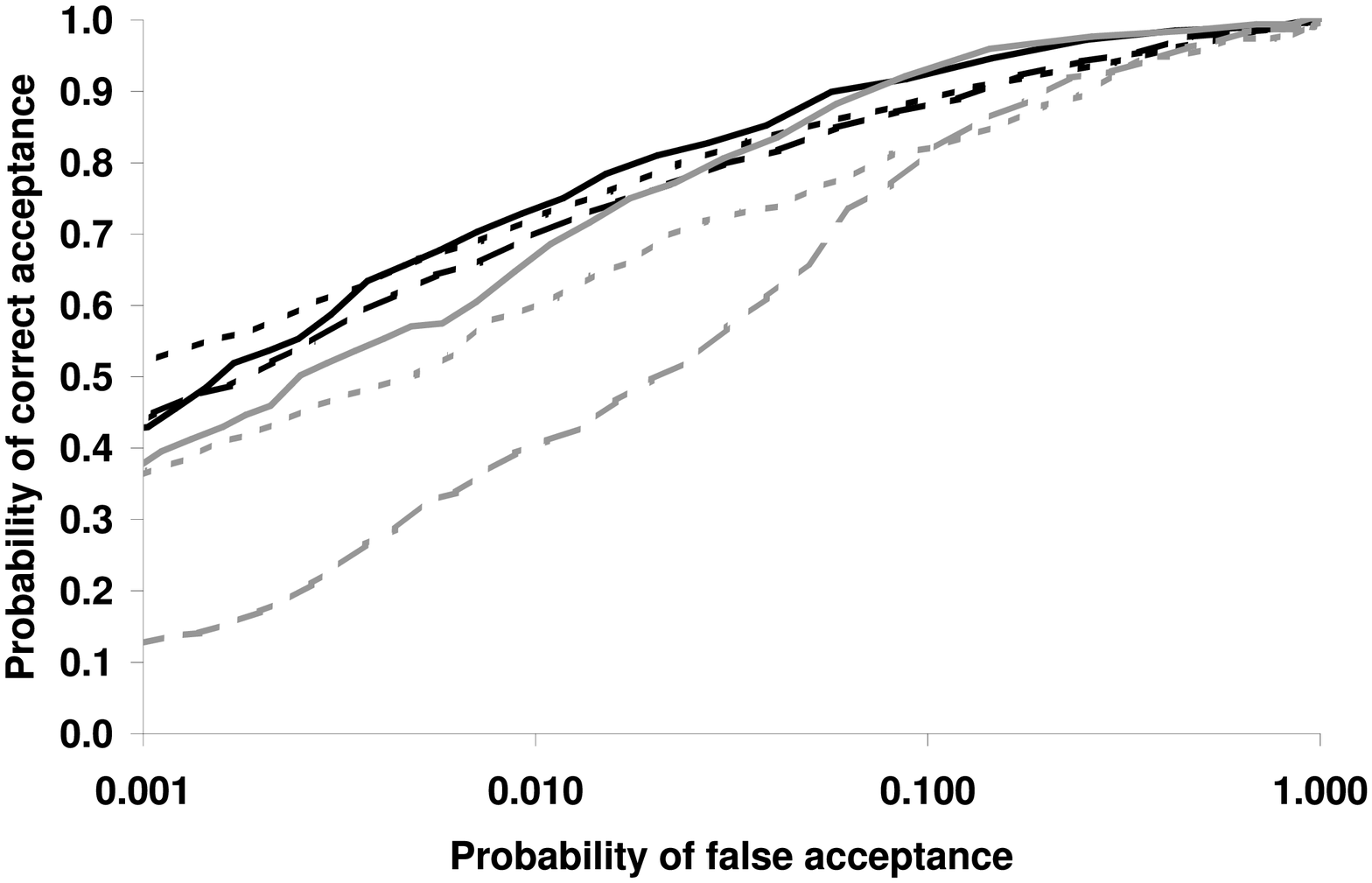, width=0.45\textwidth} \\

      Expression &
      Illumination \\
      \epsfig{file=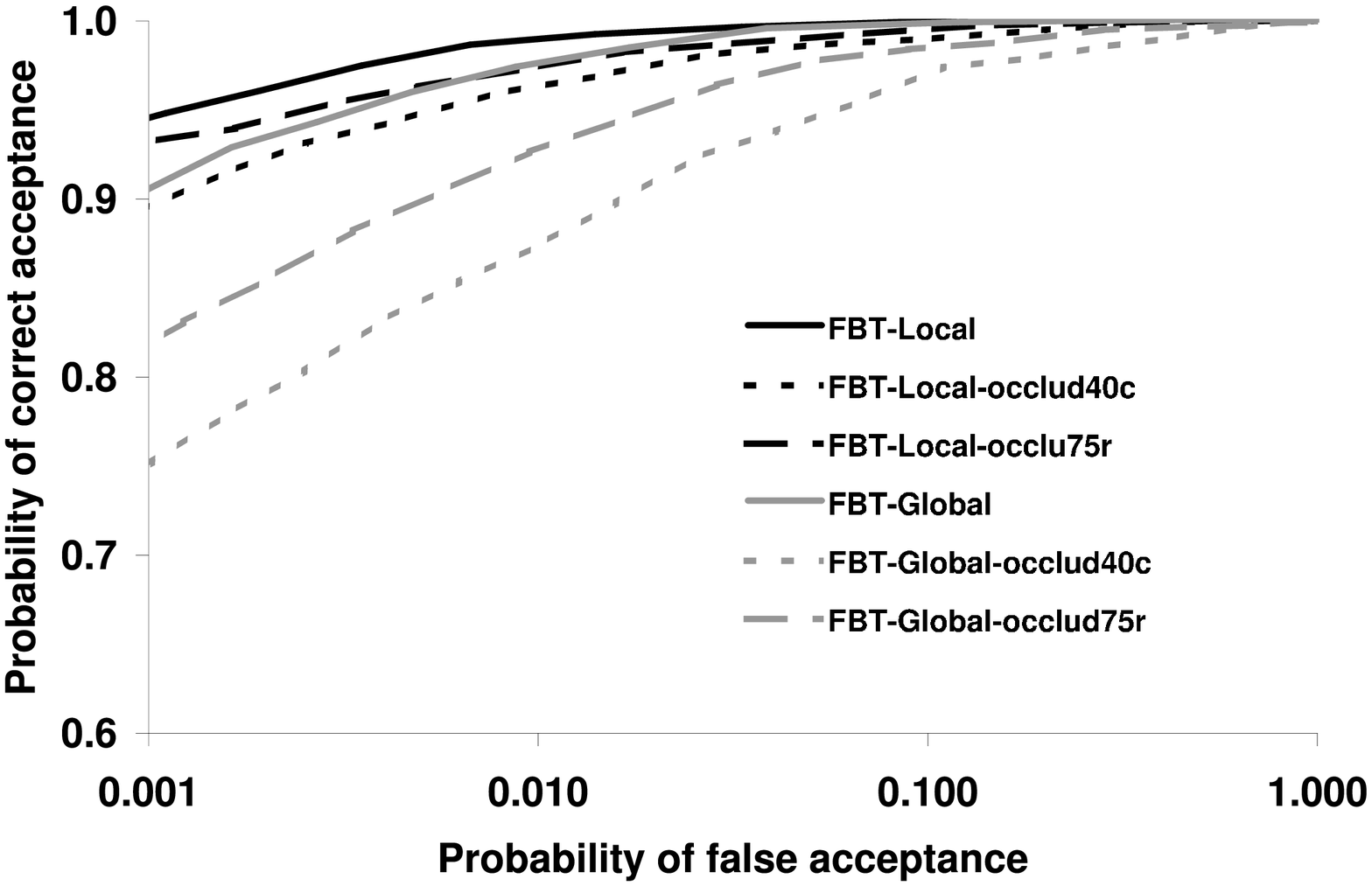, width=0.45\textwidth} &
      \epsfig{file=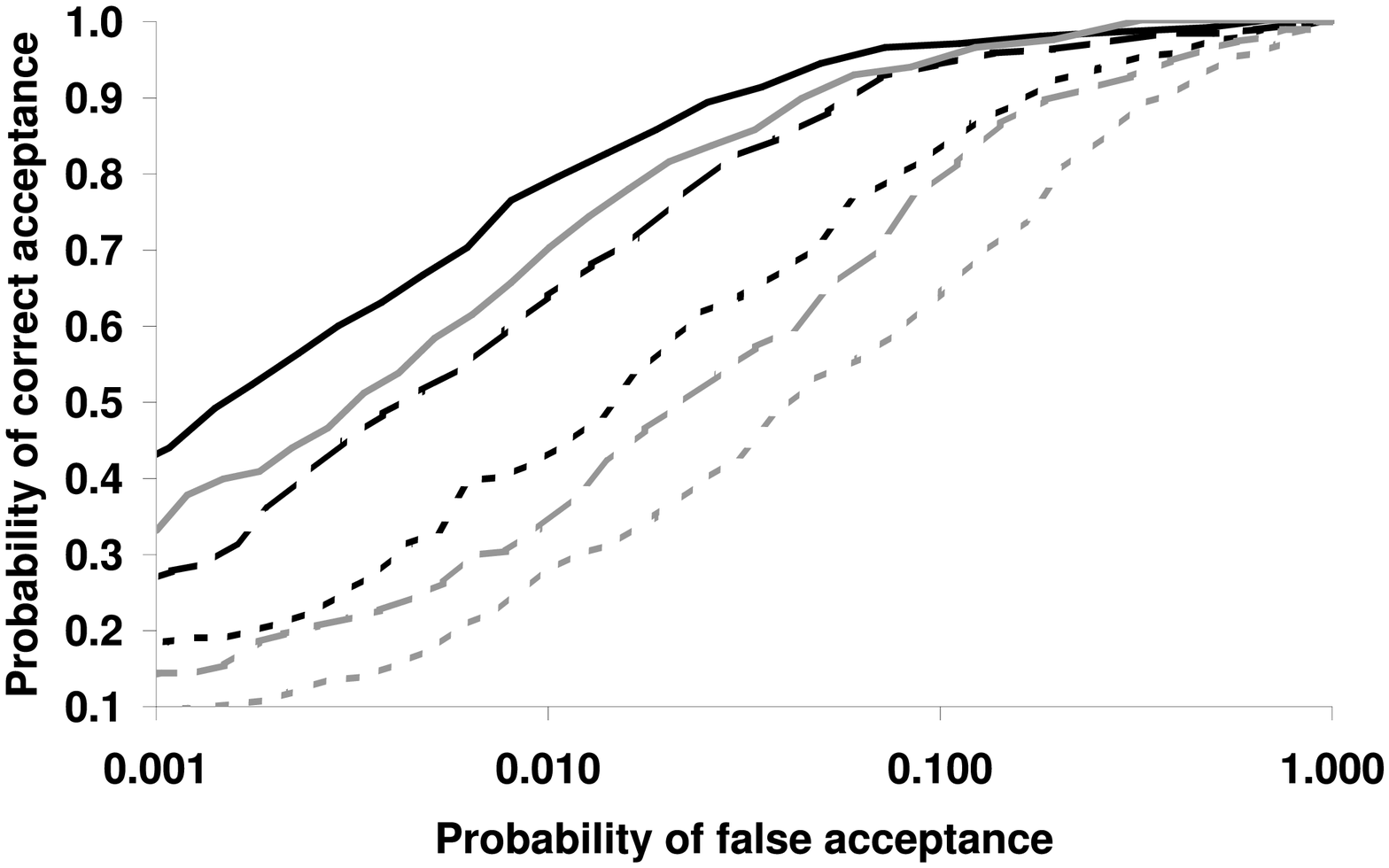, width=0.45\textwidth} \\

  \end{tabular}
  \caption{ROC functions of the FBT on the occluded and not occluded age, expression and illumination subsets.}
  \label{fig:roc-fbt-occlusion}
  \end{center}
\end{figure*}

Local approaches for face recognition are in general more robust for
occlusions (for e.g. \cite{Lanitis2004,Martinez2002}) than global
ones. To evaluate this aspect of the proposed algorithm, we occluded
all the normalized test images with a gray mask that covered $>$50\%
of the total area. We tested two masking options: masking of the
right-eye and mouth regions or masking of the mouth and nose regions
(Fig.  \ref{fig:examples-occlusion}). Figure
\ref{fig:roc-fbt-occlusion} shows the effect of occlusion on the
performance of the global and local versions of the FBT algorithms.
The severe occlusions did not reduce much the performance of the local
algorithm on the expression and age subsets, but affected
significantly performance on the illumination subset.  The global
version performed much worse under occlusion conditions on all
subsets. These results confirm the advantage of the local over the
global approach, and demonstrate the high robustness of the local-FBT
under strong occlusion conditions combined with expression and age
variations.

\subsection{Automatic system}
\begin{figure*}[ht]
  \begin{center}
  \begin{tabular}{cc}

      Age (1-34 months) &
      Age (18-34 months) \\
      \epsfig{file=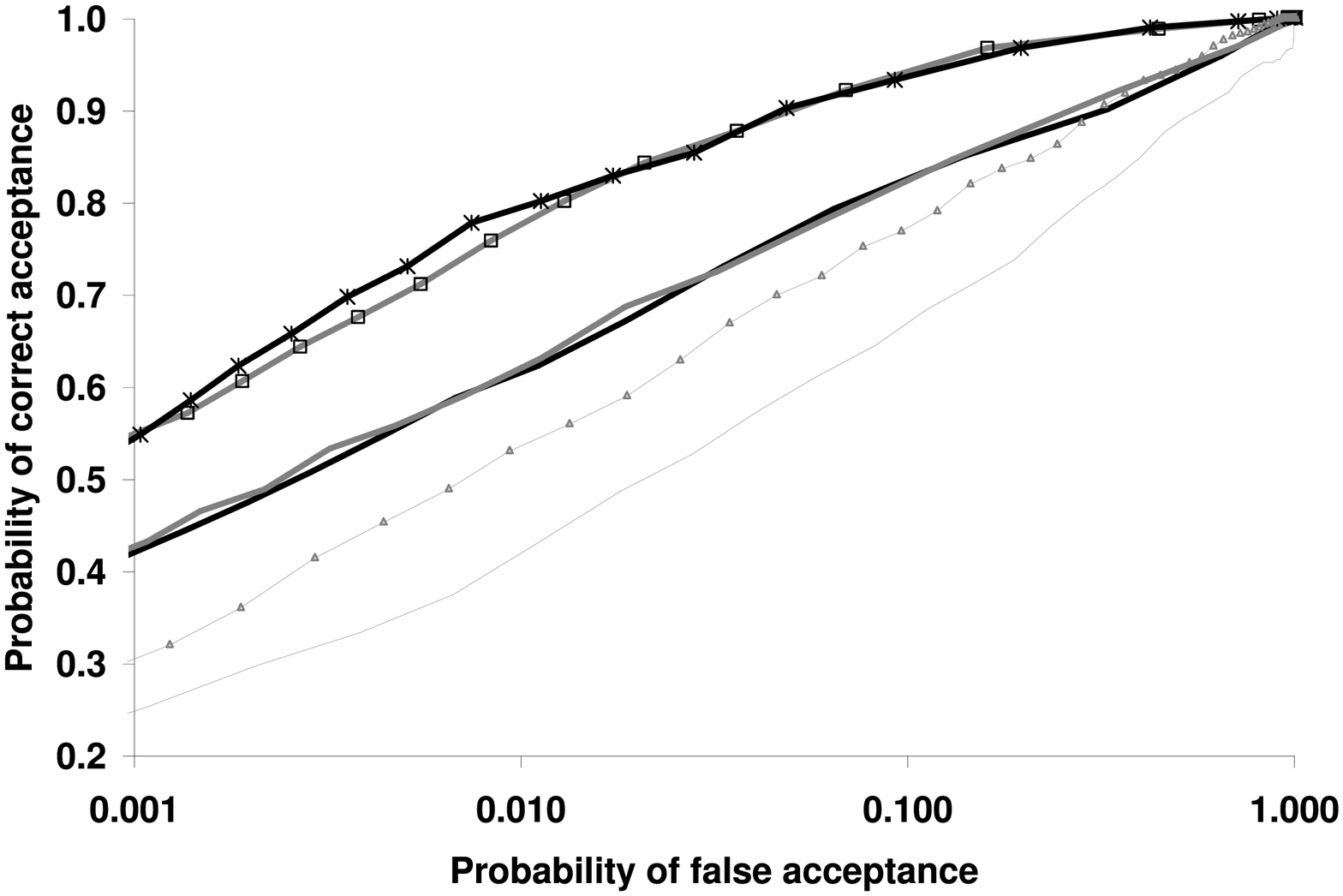, width=0.45\textwidth} &
      \epsfig{file=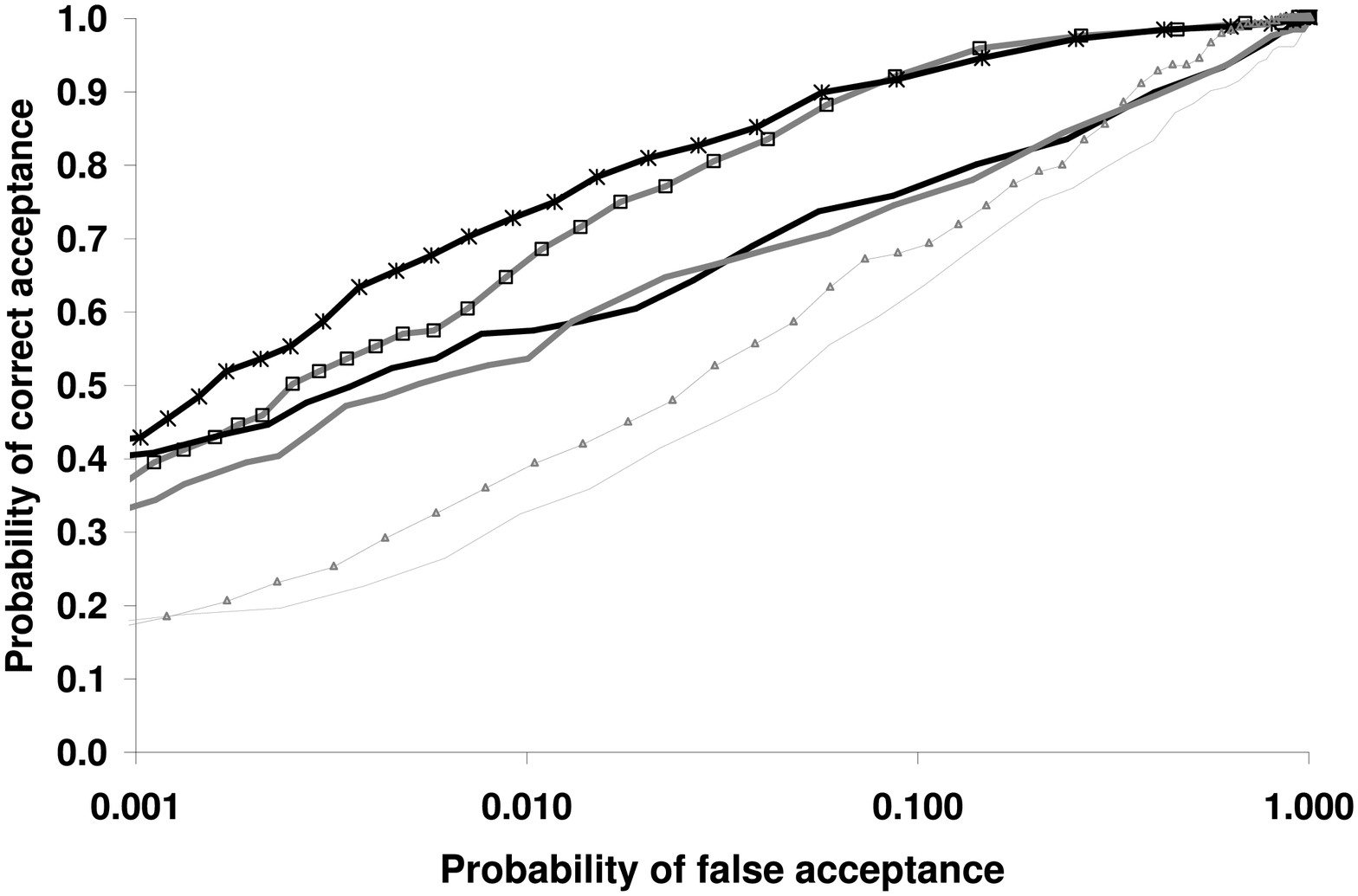, width=0.45\textwidth} \\

      Expression &
      Illumination \\
      \epsfig{file=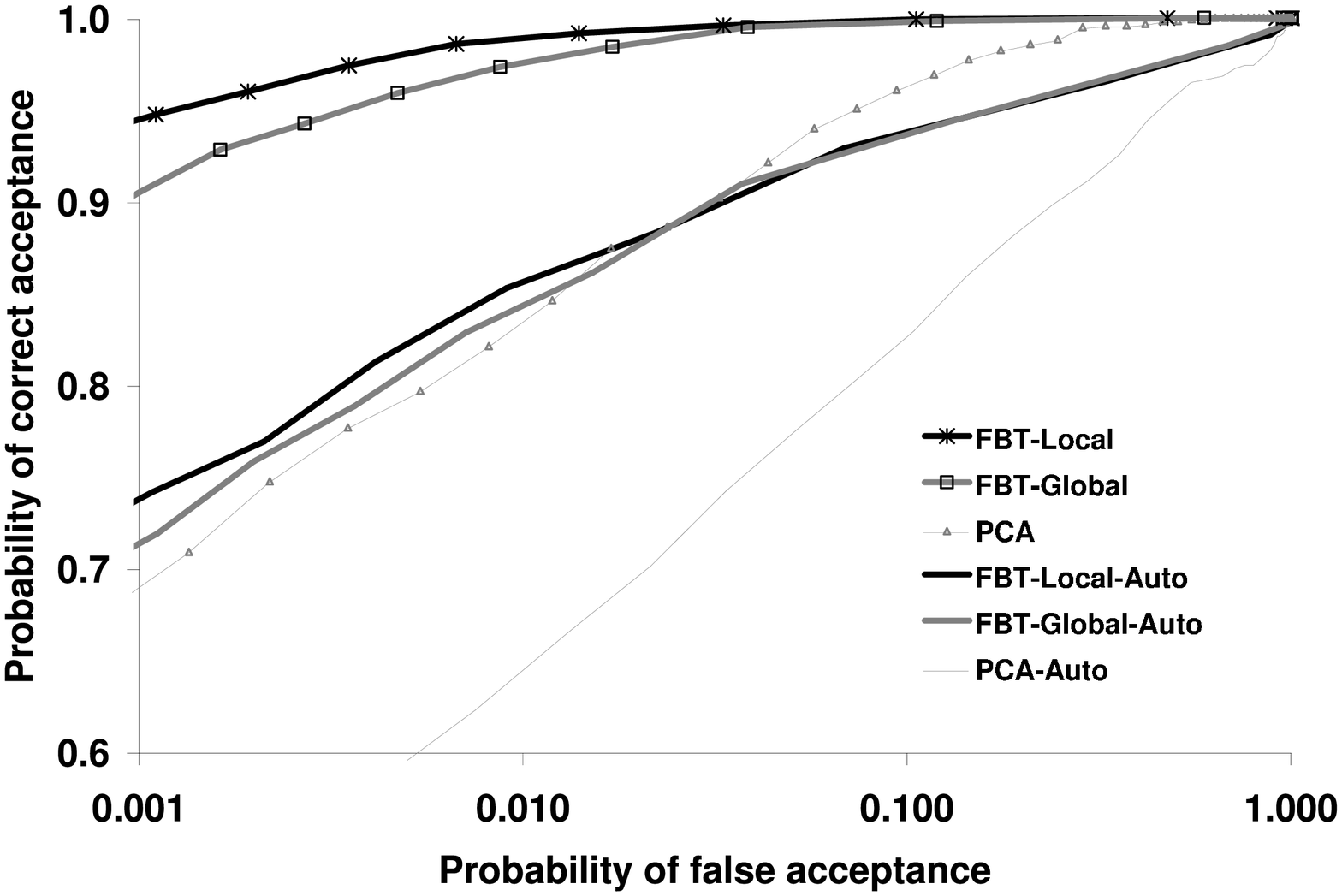, width=0.45\textwidth} &
      \epsfig{file=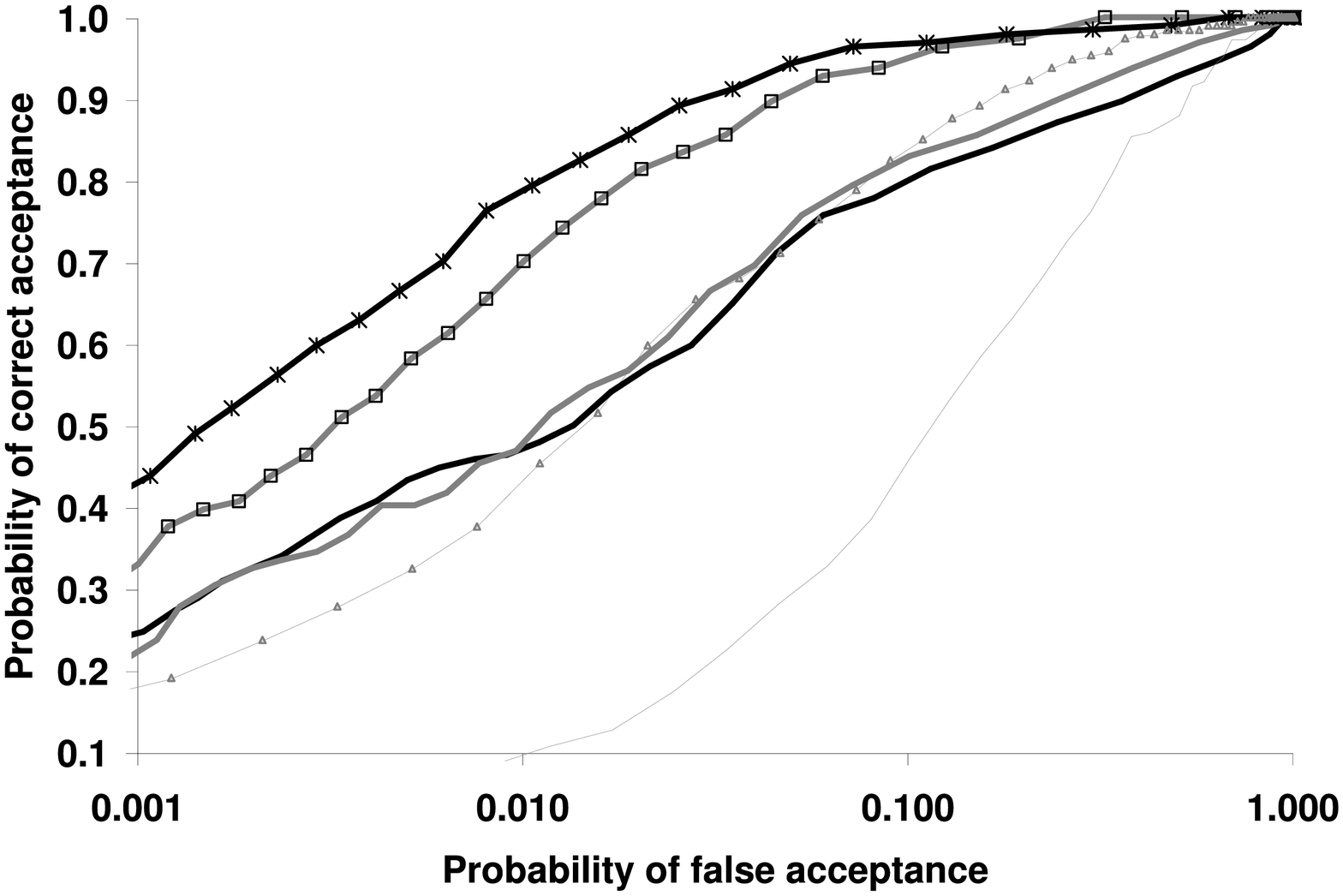, width=0.45\textwidth} \\

  \end{tabular}
  \caption{ROC functions of the semi-automatic and automatic FBT and PCA algorithms on the age, expression and illumination subsets.}
  \label{fig:roc-fbt-aam}
  \end{center}
\end{figure*}

Figure \ref{fig:roc-fbt-aam} shows the performance of the FBT
algorithms with ground-truth information and when the eyes were
detected automatically. The localization errors introduced in the
latter case reduced the performance of the FBT algorithms up to 20\%,
approximately as it affected the PCA algorithm, which is known to be
sensitive to this type of error \cite{LemieuxParizeau2002}. The
localization sensitivity of the proposed system is expected,
considering the variance property of the FBT to translation
\cite{Cabreraetal1992}. It is interesting to notice, however, that
under such conditions the advantage of the local over the global
approach was significantly reduced.

\section{Discussion}

We introduced a fully automated biologically-motivated local-based
system for face verification tasks. The main empirical result of this
study is the demonstration of the high performance of a verification
system based on FBT descriptors, especially when these are extracted
locally. The significant advantage of the FBT approach in the age
variance tests are an indication of the robustness of the polar
features in realistic situations of face variations that exceeds
simple facial expression, like illumination and age. The superior
behavior of the local approach was especially strong w.r.t.
robustness to occlusion. In the local version, the mouth region is
completely ignored, thus its occlusion or variation (ex. due to a new
beard or a scarf) does not affect performance at all.  Moreover, the
local-FBT outperformed the global-FBT even when the occluded regions
included face regions that were analyzed by the local version.

The property of robustness to occlusion of local analysis was explored
by others. Local PCA was used in \cite{Lanitis2004,Martinez2002} to
detect occluded regions in face images. The test images were
classified by comparing the unoccluded regions to corresponding
regions in training images.  However, the combination of FBT features
and local approach has several advantages over that method besides
performance.  In our proposal there, is no need to detect the occluded
regions in the test images. Furthermore, there is no need for special
training strategies \cite{Martinez2002} or for training of specific
classifiers for each testing image depending on the occluded region
\cite{Lanitis2004}. Finally, there is no need for any classification
rule for the combination of the local features; the FBT features form
a single vector. It is hard to compare our performance results with
those obtained by \cite{Lanitis2004,Martinez2002}, since their tests
were performed on subsets of less than 100 images. The training and
test images also did not included variations of expression,
illumination or age. The algorithm of \cite{Martinez2002} was adapted
in order to deal with expression variation by weighting differently
local areas and assuming that the facial expression of the training
images is known. In contrast, here we show that the proposed system
can deal simultaneously with expression, illumination, and age
variations, besides large scale occlusions.

Performance gain of the automatic FBT method can be achieved by
improving the eyes localization algorithm. For example,
\cite{Martinez2002} learned the subspace that represents localization
errors within eigenfaces. This method can be adopted for the FBT
subspace, with the advantage of the option to exclude from the final
classification face regions that gives high localization errors.

The relation of the present algorithm to human face recognition was
not directly evaluated here, but a few associations can be done. As
discussed in the Introduction, there is clear evidence that the HVS
extract global radial and angular shape information, a fact that might
look incompatible with the informative advantage of the local
information pooling showed here. However, only little is known about
the size of the global pooling area. A 1.2 visual degrees pooling area
was suggested for the detection of Glass patterns
\cite{WilsonWilkinson1998}, but the spatial locations and scale
regarding face images remain as open questions.

In the proposed system, the classifier operates in non-domain-specific
metric space whose coordinates are similarity relations. The high
performance achieved by this representation indicates that the
``real-world'' proximity relations between face images are preserved
to a good extent in the constructed internal space. It is possible
that humans also use an analogous space to represent visual objects.
This hypothesis was studied by correlating the distance between
different shape objects by objective and perceptual parameters (see
\cite{Edelman1999} for a review).  Comparison of the two measurements
is usually done by a multidimensional scaling analysis (MDS), which
projects objects as points in a two-dimensional space where the
distance between the points approximate the Euclidean distance between
the original objects. For example, in one study \cite{Rhodes1988}
objective and perceptual sorting of face images were highly
correlated, especially when the objective sorting used global
features, such as age and weight of the persons in the images. Similar
results were obtained in a neurophysiological study
\cite{YoungYamane1992} in which monkeys were presented with face
images. It was found that the MDS maps obtained from the original
images and from the response patterns of neurons in the inferotemporal
cortex had similar patterns. These results indicate that representing
images in a dissimilarity space can be analogous to human
representation mechanisms.

In conclusion, the proposed system combines high face verification
performance for expression, age, and illumination tests, and
robustness to occlusion. Future investigations, using psychophysical
methods, should establish the level of its relation to biological
systems.

\subsection*{Acknowledgments}
Y. Zana is grateful to FAPESP (03/07519-0) and to CNPq (478384/01-7).
R. Cesar-Jr. is grateful to FAPESP (99/12765-2) and to CNPq
(300722/98-2 and 474596/2004-4).  R. Barbosa is grateful to FAPESP
(03/03506-0). The authors are grateful to Rogério S. Feris and Matthew
Turk for useful comments and collaboration on this research.

\bibliographystyle{latex8}
\bibliography{bib_yossi_050413}

\end{document}